\documentclass[10pt,twocolumn,letterpaper]{article}

\usepackage{iccv}
\usepackage{epsfig}
\usepackage{times}
\usepackage{soul}
\usepackage{url}
\usepackage[utf8]{inputenc}
\usepackage[small]{caption}
\usepackage{graphicx}
\usepackage{amsmath}
\usepackage{amsthm}
\usepackage{booktabs}
\usepackage[utf8]{inputenc} 
\usepackage[T1]{fontenc}    
\usepackage{amsfonts}       
\usepackage{nicefrac}       
\usepackage{microtype}      
\usepackage{amsmath,amssymb,amsfonts}
\usepackage{subfigure}
\usepackage{subfigmat}
\usepackage{tabularx}
\usepackage{bigstrut,multirow,rotating,threeparttable}
\usepackage{enumerate}
\usepackage{textcomp}
\usepackage{xcolor}
\usepackage{float}
\usepackage{stfloats}
\usepackage{pifont}
\usepackage{caption}
\usepackage{color}
\usepackage{colortbl}
\usepackage{wrapfig}
\usepackage{graphicx}
\usepackage{eucal}
\usepackage{bm}
\usepackage{bbm}
\usepackage{algorithm,algorithmicx,algpseudocode}
\usepackage[T1]{fontenc}
\usepackage[utf8]{inputenc}
\usepackage{authblk}

\usepackage[breaklinks=true,bookmarks=false]{hyperref}

\iccvfinalcopy 

\def\iccvPaperID{****} 

\ificcvfinal\fi

\begin{document}

\title{EQ-Net: Elastic Quantization Neural Networks}

\author[1,2]{Ke Xu}
\author[4]{Lei Han}
\author[2,3]{Ye Tian}
\author[1,2*]{Shangshang Yang}
\author[1,2,3]{Xingyi Zhang}
\affil[1]{School of Artificial Intelligence, Anhui University, Hefei, China}
\affil[2]{Key Laboratory of Intelligent Computing and Signal Processing of Ministry of Education, Hefei, China}
\affil[3]{Institutes of Physical Science and Information Technology, Anhui University, Hefei, China}
\affil[4]{School of Computer Science and Technology, Anhui University, Hefei, China}

\renewcommand\Authands{ and }

\maketitle

\begin{abstract}
Current model quantization methods have shown their promising capability in reducing storage space and computation complexity. However, due to the diversity of quantization forms supported by different hardware, one limitation of existing solutions is that usually require repeated optimization for different scenarios. How to construct a model with flexible quantization forms has been less studied. In this paper, we explore a one-shot network quantization regime, named Elastic Quantization Neural Networks (EQ-Net), which aims to train a robust weight-sharing quantization supernet. First of all, we propose an elastic quantization space (including elastic bit-width, granularity, and symmetry) to adapt to various mainstream quantitative forms. Secondly, we propose the Weight Distribution Regularization Loss (WDR-Loss) and Group Progressive Guidance Loss (GPG-Loss) to bridge the inconsistency of the distribution for weights and output logits in the elastic quantization space gap. Lastly, we incorporate genetic algorithms and the proposed Conditional Quantization-Aware Accuracy Predictor (CQAP) as an estimator to quickly search mixed-precision quantized neural networks in supernet. Extensive experiments demonstrate that our EQ-Net is close to or even better than its static counterparts as well as state-of-the-art robust bit-width methods. Code can be available at \href{https://github.com/xuke225/EQ-Net.git}{https://github.com/xuke225/EQ-Net}.

\end{abstract}

\section{Introduction}
Deploying intricate deep neural networks(DNN) on edge devices with limited resources, such as smartphones or IoT devices, poses a significant challenge due to their demanding computational and memory requirements. Model quantization~\cite{QuantSurvey,QuanWhite,bnn-survey} has emerged as a highly effective strategy to mitigate the aforementioned challenge. This technique involves transforming the floating-point values into fixed-point values of lower precision, thereby reducing the memory requirements of the DNN model without altering its original architecture. Additionally, computationally expensive floating-point matrix multiplications between weights and activations can be executed more efficiently on low-precision arithmetic circuits, leading to reduced hardware costs and lower power consumption.

Despite the evident advantages in terms of power and costs, quantization incurs added noise due to the reduced precision. However, recent research has demonstrated that neural networks can withstand this noise and maintain high accuracy even when quantized to 8-bits using post-training quantization (PTQ) techniques~\cite{TensorRT,LAPQ,AdaRound,BRECQ,QDrop}. PTQ is typically efficient and only requires access to a small calibration dataset, but its effectiveness declines when applied to low-bit quantization ($\leq$ 4-bits) of neural networks. In contrast, quantization-aware training (QAT)~\cite{Dorefa,PACT,DSQ,LSQ,LSQ+,DAQ,Oscillations} has emerged as the prevailing method for achieving low-bit quantization while preserving near full-precision accuracy. By simulating the quantization operation during training or fine-tuning, the network can adapt to the quantization noise and yield better solutions than PTQ.

Currently, most AI accelerators support model quantization, but the forms of quantization supported by different hardware platforms are not exactly the same~\cite{MQBench}. For example, NVIDIA's GPU adopts channel-wise symmetric quantization in TensorRT~\cite{Tensorrt-URL} inference engine, while Qualcomm's DSP adopts per-tensor asymmetric quantization in SNPE~\cite{SNPE} inference engine. For conventional QAT methods, the different quantization forms supported by hardware platforms may require repeated optimization of the model during deployment on multiple devices, leading to extremely low efficiency of model quantization deployment.

\begin{figure*}
\centering
\includegraphics[width=1.94\columnwidth]{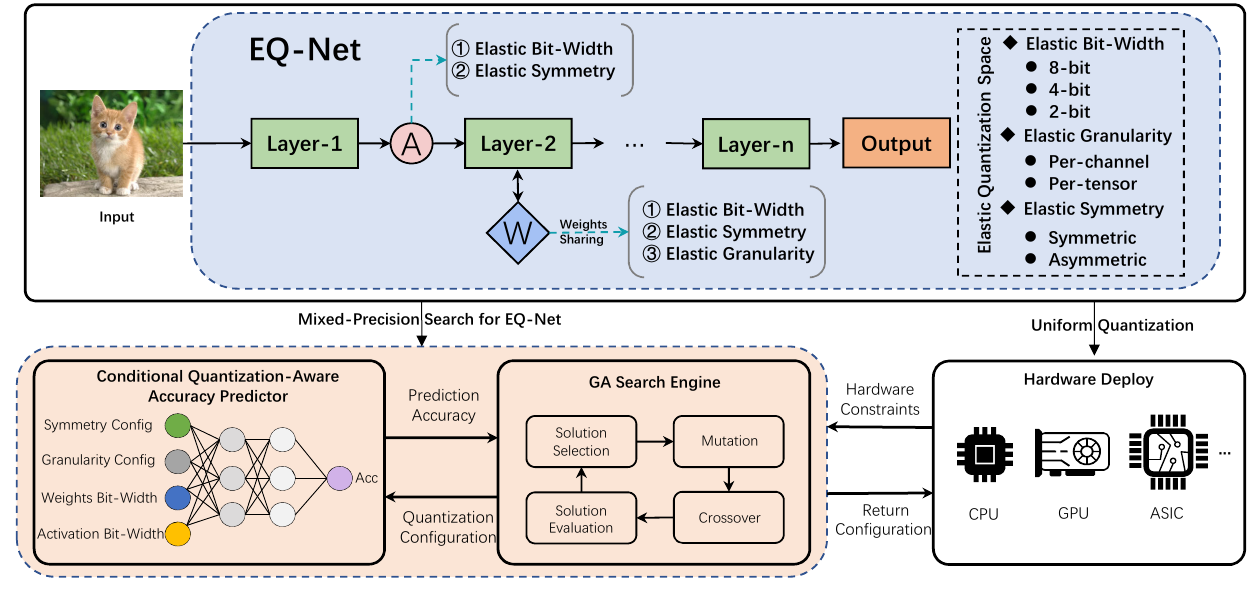}
        \caption{A conceptual overview of EQ-Net approach.} \label{fig:framework}%
\end{figure*}

To address the problem of repeated optimization in model quantization resulting from discrepancies in quantization schemes, this paper proposes an elastic quantization space design that encompasses the current mainstream quantization scenarios and classifies them into elastic quantization bit-width (2-bit, 4-bit, 8-bit, etc.), elastic quantization granularity (per-layer quantization, per-channel quantization), and elastic quantization symmetry (symmetric quantization, asymmetric quantization), as shown in Figure~\ref{fig:framework}. This approach enables flexible deployment models under different quantization scenarios by designing a unified quantization formula that integrates various model quantization forms and implementing elastic switching of quantization bit-width, granularity, and symmetry through parameter splitting.

Inspired by one-shot neural architecture search~\cite{OFA,BigNAS,AttentiveNAS,ScaleNet}, this paper attempts to train a robust elastic quantization supernet based on the constructed elastic quantization space. Unlike neural architecture search, the elastic quantization supernet is fully parameter-shared, and there is no additional weight parameter optimization space with network structure differences. Therefore, training the elastic quantization supernet may encounter the problem of negative gradient suppression~\cite{EQLoss,MultiQuant} due to different quantization forms. In other words, samples with inconsistent predictions between quantization configuration A (e.g., 8-bit/per-channel/asymmetric) and quantization configuration B (e.g., 2-bit/per-tensor/symmetric) are considered negative samples by each other, which slows down the convergence speed of the supernet during training. To solve the aforementioned problem, this paper proposes an efficient training strategy for elastic quantization supernet. Our goal is to reduce negative gradients by establishing consistency in weight and logits distributions: (1) introducing the Weight Distribution Regularization (WDR) to perform skewness and kurtosis regularization on shared weights, to better align the elastic quantization space and establish weight distribution consistency; (2) introducing the Group Progressive Guidance  (GPG) to group the quantization sub-networks and guide them with progressive soft labels during the supernet training stage to establish consistency in output logits distributions.

As shown in Figure~\ref{fig:framework}, the trained elastic quantization supernet can achieve both uniform and mixed-precision quantization (MPQ). Compared with previous MPQ works~\cite{HAQ,HMQ,HAWQ,HAWQ-v2,SDQ}, our method can specify any quantization bit-width and forms in the elastic quantization space and quickly obtain a quantized model with the corresponding accuracy. With these features, we propose a Conditional Quantization-Aware Accuracy Predictor (CQAP), combined with a genetic algorithm to efficiently search for the Pareto solution on mixed-precision quantization models under the target quantization bit-width and forms.


\section{Related Works}

\paragraph{One-Shot Network Architecture Search.}
The goal of Neural Architecture Search (NAS) is to search an optimal architecture within a large architecture search space. The term `one-shot' alludes to the fact that the subnet population only needs to be trained once. Regarding one-shot NAS methods, Cai et al.~\cite{OFA} proposed a once-for-all (OFA) model that facilitates various architectural settings by decoupling the training and search stages, thereby reducing the computational cost. BigNAS~\cite{BigNAS} challenges the conventional pipeline by training the supernet using the sandwich rule, constructing a big single-stage model without extra retraining or post-processing. AttentiveNAS~\cite{AttentiveNAS} improves the quality of the subnet by replacing the original uniform sampling strategy with a Pareto-aware sampling strategy during the training stage, and uses the Monte Carlo sampling to accelerate the sampling process. AlphaNet~\cite{AlphaNet} enhances the performance of the subnet by utilizing Alpha divergence to tackle the issue of overestimating the uncertainty of teacher networks that arise from KL divergence. Inspired by this OFA NAS approach, we construct a weight-sharing elastic quantization supernet which includes elastic quantization bit-width, symmetry, and granularity. By training an elastic quantization supernet, a variety of quantized networks with different forms can be obtained to suit different scenarios.


\paragraph{Multi-Bit Quantization of Neural Networks.}
Recently, several research works on multi-bit quantization have caught our attention. For robustness of weights, Milad et al.~\cite{l1Regularization} propose a regularization scheme applied during regular training, which models quantization noise as an additive perturbation bounded by the $\ell_\infty$ norm, constrained above the first-order term of the perturbation applied to the network from the $\ell_1$ norm of the gradients; RobustQuant~\cite{RobustQuant} prove that uniformly distributed weights have a higher tolerance to quantization with lower sensitivity to specific quantizer implementation compared to normally-distributed weights, and proposes Kurtosis regularization to enhance their quantization robustness. For robust quantization training strategies, 
AnyPrecision~\cite{AnyPrecision} employs DoReFa~\cite{Dorefa} quantization constraints to train a model but saves it in floating-point form. During runtime, the floating-point model can be directly set to different bit-widths by truncating the least significant bits; 
CoQuant~\cite{CoQuant} introduce a collaborative knowledge transfer approach to train a multi-bit quantization network;
OQAT~\cite{OQA} presents the bit inheritance mechanism under the OFA framework to progressively reduce the bit-width, allowing higher bit-width models to guide the search and training of lower bit-width models. However, this method limits its quantization policy search space to fixed-precision quantization policies, which may reduce the flexibility of the model;
BatchQuant~\cite{BatchQuant} proposes a quantizer to stabilize single-shot supernet training for joint mixed-precision quantization and architecture search;
MultiQuant~\cite{MultiQuant} enhances supernet training by using an adaptive soft label strategy to overcome the vicious competition between high bit-width and low bit-width quantized networks.
The previous studies mainly focused on the robustness of multi-bit quantization, while this paper incorporates the granularity and symmetry of quantization into the search space from the perspective of hardware deployment. In addition, by establishing similarity constraints on the weight distribution and output logits distribution, the training efficiency of the supernet is improved.

\section{Approach}
In this section, we will give a comprehensive and detailed analysis of our proposed method, mainly including the design of elastic quantization search space, the modeling of quantization supernet, and the training strategy.

\subsection{Quantization Preliminaries}
To help modeling elastic quantization neural networks, we start by introducing common notations for quantization. We introduce $\boldsymbol{w}$ and $\boldsymbol{x}$ to represent the weight matrix and activation matrix in the neural network. A complete uniform quantization process consists of quantization and de-quantization operations, which can be represented as follows:
\begin{equation} \label{eq:quant}
\left\{\begin{array}{l}
\hat{\boldsymbol{w}}=\operatorname{clip}\left(\left\lfloor\frac{\boldsymbol{w}}{s}\right\rceil+z, -2^{b-1}, 2^{b-1}-1\right) \\
\overline{\boldsymbol{w}} =s \cdot(\hat{\boldsymbol{w}}-z)
\end{array}\right.
\end{equation}
where $s$  and $z$ are called quantization step size and zero-point, respectively. $\lfloor\cdot \rceil$ rounds the continuous numbers to the nearest integers. $b$ represents the predetermined quantization bit-width. 
Given a quantization weight matrix $\hat{\boldsymbol{w}}$ and activation matrix $\hat{\boldsymbol{x}}$, the product is given by
\begin{equation} \label{eq:product}
\boldsymbol{o}_{i j} =s_w s_x \sum_{c=1}^C\left(\hat{\boldsymbol{w}}_{i c} \hat{\boldsymbol{x}}_{c j}-z_w \hat{\boldsymbol{x}}_{c j}-z_x \hat{\boldsymbol{w}}_{i c}+z_w z_x\right)
\end{equation}
where $\boldsymbol{o}$ is the convolution output or the pre-activation, $C$ represents the number of weights channels.

\subsection{Elastic Quantization Space Design}\label{section:Space}
Our elastic quantization search space consists of three parts, elastic quantization bit width, elastic quantization symmetry, and elastic quantization granularity.
\paragraph{Elastic Quantization Bit-Width.}
With proper training, different quantization bit-widths can share the same weights. Therefore, for elastic quantization bit-widths, we only need to separate and store the quantization step size and zero-point required for different quantization bit-widths. In other words, the model weights are shared among different quantization bit-widths, and only differences in quantization step size and zero-point. Typically, the quantization step size is smaller and the saturation truncation range is larger for higher bit-widths, while the quantization step size is larger and the saturation truncation range is smaller for lower bit-widths. This greatly alleviates the training pressure on hyperparameters, but poses challenges to the robustness of shared weights. Additionally, the choice of elastic quantization bit-widths is arbitrary and can be designed according to requirements.
\paragraph{Elastic Quantization Symmetry.}
Elastic quantization symmetry supports both symmetric and asymmetric quantization. For symmetric quantization, the zero-point is fixed to 0 ($z=0$), while for asymmetric quantization, the zero-point is adjustable to different ranges ($z \in \mathbb{Z}$). Asymmetric quantization scheme with trainable zero-point that can learn to accommodate the negative activations~\cite{LSQ+}. The switching between symmetric and asymmetric quantization is achieved by dynamically modifying the value of the zero point.
\paragraph{Elastic Quantization Granularity.}
Elastic quantization granularity supports both per-tensor and per-channel quantization. Per-tensor quantization uses only one set of step size and zero-point for a tensor in one layer ($s \in \mathbb{R}_{+} , z \in \mathbb{Z}$) while per-channel quantization quantizes each weight kernel independently ($s \in \mathbb{R}_{+}^{1\times C} , z \in \mathbb{Z}^{1\times C}$). Compared to per-tensor, per-channel quantization is a more fine-grained approach. Since both granularities need to be implemented in the elastic quantization space, the step size and zero-point for per-tensor can be obtained heuristically from per-channel, or can be learned as independent parameters. In addition, the elastic quantization granularity is designed for weights only, and the activations are all in the form of per-tensor.

\subsection{Elastic Quantization Network Modeling}
Assuming that the elastic quantization space of a model can be represented as $\mathcal{E}=\{\mathcal{E}_b,\mathcal{E}_g,\mathcal{E}_s\}$, where $\mathcal{E}_b$, $\mathcal{E}_g$, and $\mathcal{E}_s$ respectively represent elastic quantization bit-width, granularity, and symmetry, as described in Section~\ref{section:Space}. Given the floating-point weights $\boldsymbol{w}$ and activations $\boldsymbol{x}$, the learnable quantization step size set $\mathbf{s}=\{s_{w,l}^{e},s_{a,l}^{e} \}$, and zero-point set $\mathbf{z}=\{z_{w,l}^{e},z_{a,l}^{e} \}$, the optimization problem of the elastic quantized network can be formalized as:
\begin{equation}
\min_{\mathbf{w}^{*},\mathbf{s}^{*},\mathbf{z}^{*}} \sum_{\mathcal{E}} \mathcal{L}_{val}\left(\text{QNN}\left(\hat{\boldsymbol{w}}, \hat{\boldsymbol{x}},\boldsymbol{s},\boldsymbol{z} \right)\right)
\end{equation}
where $s_{w,l}^{e}$ and $s_{a,l}^{e}$ represent the weights and activation step size with quantization configuration $e \in \mathcal{E}$ in layer $l$; $\mathcal{L}_{val}$ denotes the validation loss; QNN denotes quantization neural network. It can be seen that the training objective of elastic quantization networks is to minimize the task loss under all elastic quantization spaces by optimizing the weights, step sizes, and zero-points.


\subsection{Elastic Quantization Training} \label{training strategy}
To enable efficient elastic quantization training, we propose the use of weight distribution regularization and group progressive guidance techniques to promote data consistency across various elastic quantization spaces.

\paragraph{Weight Distribution Regularization.} 
DNN weights often conform to Gaussian or Laplace distributions~\cite{WeightsDistribution}. To better align these weights to the elastic quantization space, we propose the incorporation of skewness and kurtosis regularizations. Skewness regularization primarily limits the direction and degree of skewness in the data distribution (as expressed in Eq.(\ref{eq:Skew}), where $\mu$ and $\sigma$ are the mean and standard deviation of $\boldsymbol{w}$). Reducing the degree of skewness in the weight distribution enhances the robustness of weights in elastic quantization symmetry.
\begin{equation}\label{eq:Skew}
\operatorname{Skew}[\boldsymbol{w}]=\mathbb{E}\left[\left(\frac{\boldsymbol{w}-\mu}{\sigma}\right)^3\right]
\end{equation}
In contrast, kurtosis regularization primarily limits the sharpness of the peak in the data distribution (as expressed in Eq.(\ref{eq:Kurt})). Reducing the sharpness of the weight distribution peak enhances the robustness of weights in the elastic quantization bit-width.
\begin{equation}\label{eq:Kurt}
\operatorname{Kurt}[\boldsymbol{w}]=\mathbb{E}\left[\left(\frac{\boldsymbol{w}-\mu}{\sigma}\right)^4\right]
\end{equation}
To sum up, the weight distribution regularization loss for the supernet training is defined as follows:
\begin{equation}
\mathcal{L}_{\text{WDR}}=\frac{1}{L} \sum_{i=1}^L\left(\left|\operatorname{Skew}\left[\boldsymbol{w}_i\right]\right|^2 + \left|\operatorname{Kurt}\left[\boldsymbol{w}_i\right]-\mathcal{K}_T\right|^2\right)
\end{equation}
where $L$ is the number of layers and $\mathcal{K}_T$ is the target for kurtosis regularization. Based on relevant experimental research~\cite{RobustQuant}, optimal robustness is achieved at $\mathcal{K}_T=1.8$.

\paragraph{Group Progressive Guidance.}  
As highlighted in~\cite{KD,KDSurvey}, an ensemble of teacher networks can provide more diverse soft labels during distillation training of the student network, leading to greater consistency in output logits. In our supernet, a multitude of subnets exists with varying quantization configurations, thereby enabling the generation of diverse soft labels. Motivated by this, we employ different grouped subnets as a teacher ensemble during in-place distillation to achieve progressive guidance across different groups. Following the sandwich rule~\cite{BigNAS}, we sample the highest quantization bit-width subnets (including random symmetry and granularity, denoted as $H$), the lowest (denoted as $L$), and random subnets (denoted as $R$) in each training step. In this approach, the subnets with the highest bit-width are trained to predict the ground truth label $\boldsymbol{y}$, while the subnets with random bit-width losses are defined based on the cross-entropy with the ground truth label and the Kullback-Leibler (KL) divergence with the soft logits of highest subnets, $\mathcal{Y}_H$. Likewise, the losses of the lowest subnets are defined based on the cross-entropy with $\boldsymbol{y}$ and the KL divergence with $\mathcal{Y}_R$. 
\begin{equation}
\left\{\begin{array}{l}
\mathcal{L}_H =\mathcal{L}_{\text{CE}}\left(\mathcal{Y}_H, \boldsymbol{y}\right) \\
\mathcal{L}_R =  \lambda * \mathcal{L}_{\text{KL}}\left(\mathcal{Y}_R , \mathcal{Y}_H \right) 
                        + (1-\lambda) * \mathcal{L}_{\text{CE}}\left(\mathcal{Y}_R , \boldsymbol{y}\right) \\
\mathcal{L}_L = \lambda * \mathcal{L}_{\text{KL}}\left(\mathcal{Y}_L, \mathcal{Y}_R\right) 
                        + (1-\lambda) * \mathcal{L}_{\text{CE}}\left(\mathcal{Y}_L, \boldsymbol{y}\right) 
\end{array}\right.
\end{equation}
where $\mathcal{L}_{\text{KL}}$ and $\mathcal{L}_{\text{CE}}$ indicate the KL divergence loss and cross-entropy loss, respectively. In summary, the group progressive guidance losses for training the supernet are defined as follows:
\begin{equation}
\mathcal{L}_{\text{GPG}}(\theta)=\mathcal{L}_H(\theta) + \mathcal{L}_R(\theta) + \mathcal{L}_L(\theta)
\end{equation}
It then aggregates the gradients from all sampled subnets before updating the weights of the supernet model. 

\subsection{Mixed-Precision Quantization Search} \label{search}
The mixed-precision search approach is designed to systematically explore the suitable bit-width configuration for each layer of a supernet. During the performance estimation phase, it is necessary to perform batch norm calibration~\cite{EagleEye,BigNAS} to re-calibrate the statistics of the batch normalization layer prior to estimating the performance of the quantization subnet. Batch norm calibration and the validation of quantization models are time-consuming, resulting in an expensive evaluation cost for the search. When employing search algorithms for quantized bit-width search, thousands of subnets must be evaluated. To expedite the search process and minimize the time cost in the search phase, we propose a proxy model for performance estimation.

\paragraph{Conditional Quantization-Aware Accuracy Predictor.}
In the stage of mixed precision quantization, not only the bit-width of each layer but also the form of quantization will have a crucial impact on the final results. To achieve a unified prediction of the elastic quantization model, we propose a Conditional Quantization-Aware Accuracy Predictor (CQAP) in contrast to previous precision predictors~\cite{MultiQuant}. As shown in the lower left corner of Figure~\ref{fig:framework}, we use the quantization symmetry and granularity as the conditions to evaluate the final precision for different bit-widths, and adopt binary encoding as the input to the predictor. The backbone architecture of the predictor maintains the same MLP structure as the previous work~\cite{AttentiveNAS,MultiQuant}, and the output results in the predicted accuracy. The CQAP can be formalized as:
\begin{equation}
 \text{acc}= \text{MLP}(\underbrace{G_w,S_w,S_a}_{\mathrm{Conditional}},\underbrace{B_w,B_a}_{\mathrm{BitWidth}})
\end{equation}
where $G_w$, $S_w$, $B_w$ represent the granularity, symmetry, and bit width of each layer for weights quantization respectively. $S_a$, $B_a$ represent the symmetry and bit width of each layer for activations quantization respectively.

\paragraph{Genetic Algorithm for Mixed-Precision Search.}
During the search phase, the genetic algorithm\cite{whitley1994genetic} explores the bit-width of each layer and utilizes a CQAP to evaluate the corresponding accuracy of each candidate configuration. The genetic algorithm first initializes a set of solutions that satisfy the constraints using Monte Carlo sampling~\cite{MultiQuant,AlphaNet} as the initial population. Subsequently, the fitness score of each candidate quantization network produced by the predictor is evaluated based on its accuracy. The individual with the highest fitness scores is preserved as elitist and included in the mutation and crossover process to generate a new population based on a predefined probability.  This selection-mutation-crossover procedure is iteratively performed until the algorithm achieves a satisfactory Pareto solution that satisfies the average bit-width targets for both weights and activations.

\section{Experimental Results} \label{Experimental}
In this section, we present the results of a comprehensive set of experiments demonstrating the superiority of our proposed approach over several baselines on the ImageNet~\cite{imagenet}. Additionally, we conducted comprehensive ablation experiments and visualization analyses to confirm the effectiveness of both the WDR and the GPG methods for EQ-Net.

\subsection{Implementation Details} \label{Implementation Details}
We separately trained two major classes of models using pre-trained weights provided by the TorchVision and PyTorch v1.10 frameworks~\cite{pytorch}. The first class comprised classical ResNet~\cite{ResNet} models, namely ResNet18 and ResNet50, while the second class included lightweight models MobileNetV2~\cite{MobileNetV2} and EfficientNetB0~\cite{EfficientNet}, which utilize separable convolutions.
It is worth mentioning that the EfficientNetB0 model utilizes the Swish~\cite{Swish} activation function, which produces negative values. This feature allows us to investigate the differences between symmetric and asymmetric quantization using this model.
The elastic quantization space of these networks is shown in Table~\ref{tab:space}. 
Note that we excluded 2-bit quantization in the lightweight model, as it results in a significant performance drop.
We train each model for 120 epochs using Adam~\cite{Adam} optimizer with a cosine learning rate decay. The base learning rate is set as 0.001. 
After each quantization supernet is trained, we sample 8000 different subnetworks in each supernet and calculate their accuracy on a subset of the training set, making a <config, accuracy> dataset to train CQAP. We train CQAP for 100 epochs using SGD, the learning rate is set as 0.0004, and the weight decay of 0.0001. In the search phase of GA, we set the size of the population to 100 and the number of generations to 500.
 
\begin{table*}[htbp]
  \centering
  \caption{Elastic quantization space design under different models}
  \scalebox{0.88}{
    \begin{tabular}{|c|c|c|c|c|c|}
    \hline
    \multirow{2}[4]{*}{NetWork} & \multicolumn{3}{c|}{Weight Quantization Forms} & \multicolumn{2}{c|}{Activation Quantization Forms} \bigstrut\\
\cline{2-6}          & Bit-Width & Symmetric & Granularity & Bit-Width & Symmetric \bigstrut\\
    \hline
    ResNet18/ResNet50 & 2,3,4,5,6,7,8 & symmetric/asymmetric & per-channel/per-layer & 2,3,4,5,6,7,8 & symmetric/asymmetric \bigstrut\\
    \hline
    MobileNetV2/EfficientNetB0 & 3,4,5,6,7,8 & symmetric/asymmetric & per-channel/per-layer & 3,4,5,6,7,8 & symmetric/asymmetric \bigstrut\\
    \hline
    \end{tabular}%
    }
  \label{tab:space}%
\end{table*}%

\begin{table*}[htbp]
  \centering
  \caption{Comparison of state-of-the-art quantization methods on ImageNet. `B-OFA' denotes bit-width One-For-All methods, `BGS-OFA' denotes bit-width, symmetry and granularity One-For-All methods.}
  \scalebox{0.735}{
    \begin{tabular}{ccccccccccc}
    \hline
    \multirow{2}[2]{*}{\textbf{Network}} & \multirow{2}[2]{*}{\textbf{Benchmark}} & \multirow{2}[2]{*}{\textbf{Criterion}} & \multirow{2}[2]{*}{\textbf{Granularity}} & \multirow{2}[2]{*}{\textbf{ Symmetry}} & \multicolumn{2}{c}{\textbf{Weights}} & \multicolumn{2}{c}{\textbf{Activation}} & \multicolumn{2}{c}{\textbf{Accuracy}} \bigstrut[t]\\
          &       &       &       &       & \textbf{W-bits} & \textbf{W-Comp} & \textbf{A-bits} & \textbf{A-Comp} & \textbf{Top-1 (Drop)} & \textbf{FP Top-1} \bigstrut[b]\\
    \hline
    \multirow{11}[6]{*}{ResNet-18} & LSQ~\cite{LSQ}   & Uniform & Per-tensor & Symmetric & 2     & 14.11$\times$ & 2     & 13.25$\times$ & \textbf{67.6\% (↓2.9\%)} & 70.5\% \bigstrut[t]\\
          & LSQ+~\cite{LSQ+}  & Uniform & Per-tensor & Asymmetric & 2     & 14.11$\times$ & 2     & 13.25$\times$ & 66.8\% (↓3.3\%) & 70.1\% \\
          & EdMIPS~\cite{EdMIPS} &  Mixed-Precision & Per-tensor & Symmetric & 2 MP  & 16.00$\times$ & ——    & <16.00$\times$ & 65.9\% (↓3.9\%) & 69.8\% \bigstrut[b]\\
\cline{2-11}          & RobustQuant~\cite{RobustQuant} & B-OFA & Per-tensor & Symmetric & 3     & 10.67$\times$  & 3     & 10.67$\times$  & 57.3\% (↓13.0\%) & 70.3\% \bigstrut[t]\\
          & CoQuant~\cite{CoQuant} & B-OFA & Per-tensor & Symmetric & 2     & 14.11$\times$  & 2     & 13.25$\times$  & 57.1\% (↓12.7\%) & 69.8\% \\
          & AnyPrecision~\cite{AnyPrecision} & B-OFA & Per-tensor & Symmetric & 2     & 14.11$\times$  & 2     & 13.25$\times$  & 64.2\% (↓4.0\%) & 68.2\% \\
          & MultiQuant~\cite{MultiQuant} & B-OFA & Per-tensor & Asymmetric & 3     & 10.37$\times$  & 3     & 10.37$\times$  & 67.5\% (↓2.3\%) & 69.8\% \\
          & MultiQuant~\cite{MultiQuant} & B-OFA & Per-tensor & Asymmetric & 3 MP  & 9.93$\times$  & 3 MP  & 9.56$\times$  & 69.2\% (↓0.6\%) & 69.8\% \bigstrut[b]\\
\cline{2-11}          & \multirow{3}[2]{*}{EQ-Net(Ours)} & \multirow{3}[2]{*}{BGS-OFA } & Per-tensor & Symmetric & 2     & 14.11$\times$  & 2     & 13.25$\times$  & \textbf{65.9\% (↓3.9\%)} & 69.8\% \bigstrut[t]\\
          &       &       & Per-tensor & Asymmetric & 3     & 10.37$\times$  & 3     & 10.37$\times$  & \textbf{69.3\% (↓0.5\%)} & 69.8\% \\
          &       &       & Per-tensor & Asymmetric & 3 MP  & 9.93$\times$  & 3 MP  & 9.56$\times$  & \textbf{69.8\% (↓0.0\%)} & 69.8\% \bigstrut[b]\\
    \hline
    \multirow{10}[6]{*}{ResNet-50} & LSQ~\cite{LSQ}   & Uniform & Per-tensor & Symmetric & 2     & 12.88$\times$  & 2     & 15.34$\times$  & \textbf{73.7\% (↓3.2\%)} & 76.9\% \bigstrut[t]\\
          & HAQ~\cite{HAQ}   &  Mixed-Precision & Per-tensor & Symmetric & 3 MP  & 10.57$\times$  & MP    & ——    & \textbf{75.3\% (↓0.8\%)} & 76.1\% \\
          & HAWQ-V2~\cite{HAWQ-v2} &  Mixed-Precision & Per-channel & Symmetric & 2 MP  & 12.24$\times$  & 4 MP  & <8.00$\times$  & 75.8\% (↓1.6\%) & 77.4\% \bigstrut[b]\\
\cline{2-11}          & RobustQuant~\cite{RobustQuant} & B-OFA & Per-tensor & Symmetric & 3     & 10.67$\times$  & 3     & 10.67$\times$  & 57.3\% (↓19.0\%) & 76.3\% \bigstrut[t]\\
          & CoQuant~\cite{CoQuant} & B-OFA & Per-tensor & Symmetric & 2     & 12.88$\times$  & 2     & 15.34$\times$  & 57.1\% (↓19.0\%) & 76.1\% \\
          & AnyPrecision~\cite{AnyPrecision} & B-OFA & Per-tensor & Symmetric & 2     & 12.88$\times$  & 2     & 15.34$\times$  & 71.7\% (↓3.3\%) & 75.0\% \\
          & MultiQuant~\cite{MultiQuant} & B-OFA & Per-tensor & Asymmetric & 3     & 10.67$\times$  & 3     & 10.67$\times$  & \textbf{75.4\% (↓0.7\%)} & 76.1\% \bigstrut[b]\\
\cline{2-11}          & \multirow{3}[2]{*}{EQ-Net(Ours)} & \multirow{3}[2]{*}{BGS-OFA } & Per-tensor & Symmetric & 2     & 12.88$\times$  & 2     & 15.34$\times$  &  \textbf{72.5\%(↓3.6\%)}     & 76.1\% \bigstrut[t]\\
          &       &       & Per-tensor & Asymmetric & 3     & 10.67$\times$  & 3     & 10.67$\times$  &   \textbf{74.7\%(↓1.4\%)}    & 76.1\% \\
          &       &       & Per-tensor & Symmetric & 3 MP  & 10.57$\times$  & 3 MP  & 10.57$\times$  &   \textbf{75.1\%(↓1.0\%)}    & 76.1\% \bigstrut[b]\\
    \hline
    \multirow{5}[6]{*}{MobileNetV2} & HAQ~\cite{HAQ}   &  Mixed-Precision & Per-tensor & Symmetric & 4 MP  & 8.00$\times$  & 4 MP  & 8.00$\times$  & 67.0\% (↓5.1\%) & 72.1\% \bigstrut\\
\cline{2-11}          & RobustQuant~\cite{RobustQuant}   & B-OFA & Per-tensor & Symmetric & 4     & 8.00$\times$  & 4     & 8.00$\times$  & 59.0\% (↓12.3\%) & 71.3\% \bigstrut[t]\\
          & MultiQuant~\cite{MultiQuant} & B-OFA & Per-tensor & Asymmetric & 4     & 8.00$\times$  & 4     & 8.00$\times$  & 69.9\% (↓2.0\%) & 71.9\% \bigstrut[b]\\
\cline{2-11}          & \multirow{2}[2]{*}{EQ-Net(Ours)} & \multirow{2}[2]{*}{BGS-OFA } & Per-tensor & Asymmetric & 4     & 8.00$\times$  & 4     & 8.00$\times$  & \textbf{71.0\% (↓0.9\%)} & 71.9\% \bigstrut[t]\\
          &       &       & Per-tensor & Symmetric & 4 MP  & 8.00$\times$  & 4 MP  & 8.00$\times$  & \textbf{71.2\% (↓0.7\%)} & 71.9\% \bigstrut[b]\\
    \hline
    \multirow{4}[4]{*}{EfficientNetB0} & LSQ~\cite{LSQ}   & Uniform & Per-tensor & Symmetric & 4     & 8.00$\times$  & 4     & 8.00$\times$  & 71.9\% (↓4.2\%) & 76.1\% \bigstrut[t]\\
          & LSQ+~\cite{LSQ+}  & Uniform & Per-tensor & Asymmetric & 4     & 8.00$\times$  & 4     & 8.00$\times$  & 73.8\% (↓2.3\%) & 76.1\% \bigstrut[b]\\
\cline{2-11}          & \multirow{2}[2]{*}{EQ-Net(Ours)} & \multirow{2}[2]{*}{BGS-OFA } & Per-tensor & Symmetric & 4     & 8.00$\times$  & 4     & 8.00$\times$  &   \textbf{74.1\% (↓3.6\%)}     & 77.7\% \bigstrut[t]\\
          &       &       & Per-tensor & Asymmetric & 4     & 8.00$\times$  & 4     & 8.00$\times$  &   \textbf{75.1\% (↓2.6\%)}    & 77.7\% \bigstrut[b]\\
    \hline
    \end{tabular}%
    }
  \label{tab:SOTA}%
\end{table*}%

\subsection{Comparison with State-of-the-Art Methods} \label{SOTA}
Table~\ref{tab:SOTA} shows the comparison of our trained EQ-Net which uses Bit-width, Granularity, and Symmetry One-For-All(BGS-OFA) method with fixed quantization, mixed precision, and other Bit-width One-For-All(B-OFA) methods. 

For ResNet18, EQ-Net outperforms RobustQuant~\cite{RobustQuant} and CoQuant~\cite{CoQuant}, by nearly 10\% at 2 and 3 fixed bit-width, and this gap is further widened to 15\% in ResNet50.
When the quantization bit width is set to 3, we outperform MultiQuant~\cite{MultiQuant} by 1.8\% in ResNet18 but underperform this algorithm by 0.7\% in ResNet50. We speculate that the reason for this difference is that our BGS-OFA method contains per-channel quantization form, which is more unstable~\cite{DAQ} when the model is larger and affects the training of the whole supernet.
Compared with LSQ method, we have less than 1\% accuracy gap in the 2-bit quantization of ResNet model, but our method has better robustness and generality. In mixed precision quantization, our 3-bit mixed quantization accuracy in ResNet18 has reached the accuracy of FP32, which benefits from robust supernet training and search technology.

In both the lightweight MobileNetV2 and EfficientNetB0 models, the capability of our algorithm is further illustrated. 
In MobileNetV2, we surpass the algorithms RobustQuant and MultiQuant which use the B-OFA approach by 11.4\% and 1.1\% at 4 bit-width, respectively. Meanwhile, our algorithm outperforms HAQ~\cite{HAQ} by 4.2\% in mixed precision quantization. The reason for achieving such well-done results is that when using separable convolution, the distribution of weights in some layers is irregular and sometimes even double-peaked~\cite{BiasTuning}, increasing the difficulty of quantization, while our WDR-Loss can well transition the weights to uniform distribution and improve the accuracy of quantization.
Since the activation function used by ResNet18, ResNet50, and MobileNetV2 is ReLU~\cite{ReLU}, which has no negative values, there is not much difference between symmetric and asymmetric quantization. EfficientNetB0 uses the Swich~\cite{Swish} activation function with negative values, and we can see an improvement of about 1\% when applying asymmetric quantization compared to symmetric quantization. 
Our algorithm outperforms LSQ by 0.6\% in symmetric quantization but falls short of LSQ+~\cite{LSQ+} by 0.3\% in asymmetric quantization. This disparity can be attributed to the fact that the network weights need to balance the trade-offs between the two quantization methods, resulting in an increase in the accuracy of symmetric quantization while a little decrease in the accuracy of asymmetric quantization.

\subsection{Ablation Studies} \label{Ablation}

\begin{figure*}[!htb]
\subfigure[Baseline.] { \label{fig:baseline}
\includegraphics[width=0.51\columnwidth]{ 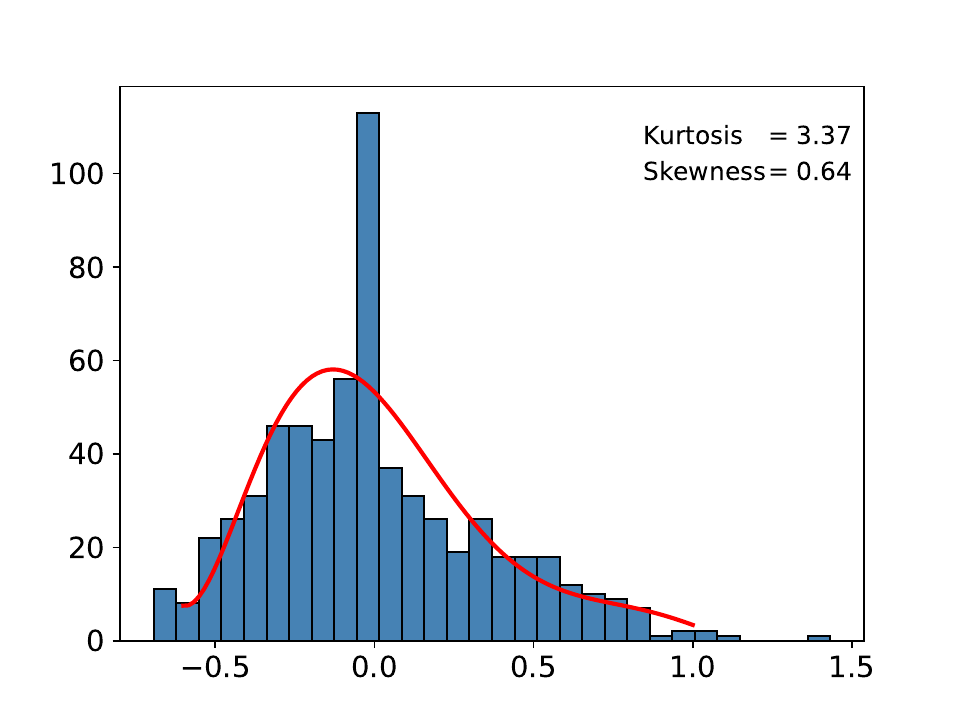} 
}
\subfigure[Kurtosis Loss.] { \label{fig:Kurtosis}
\includegraphics[width=0.492\columnwidth]{ 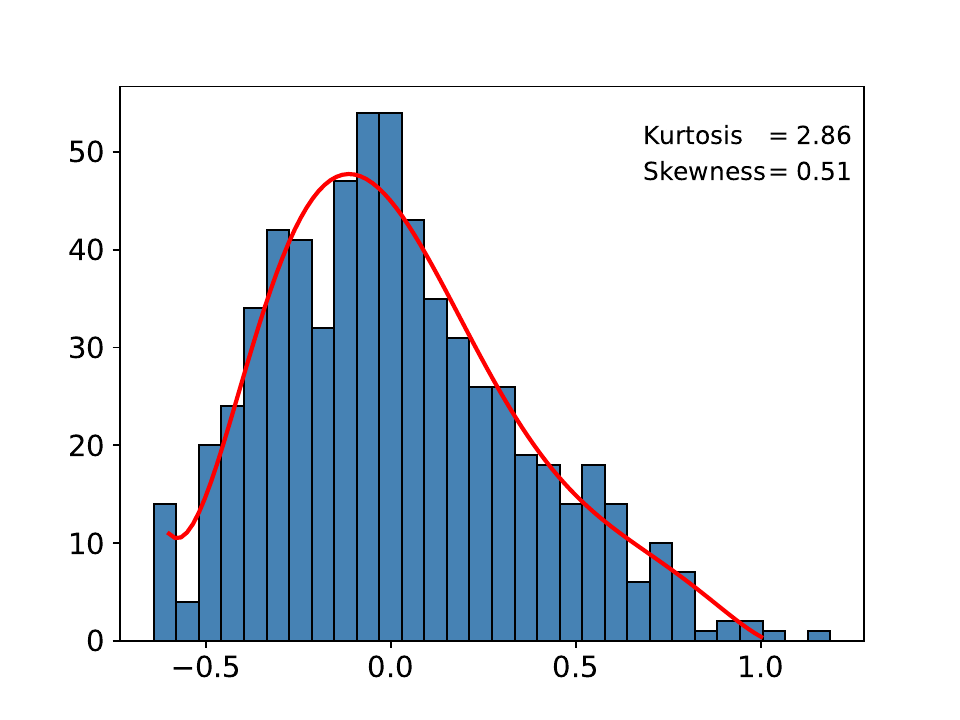} 
}
\subfigure[Skewness Loss.] { \label{fig:Skewness}
\includegraphics[width=0.492\columnwidth]{ 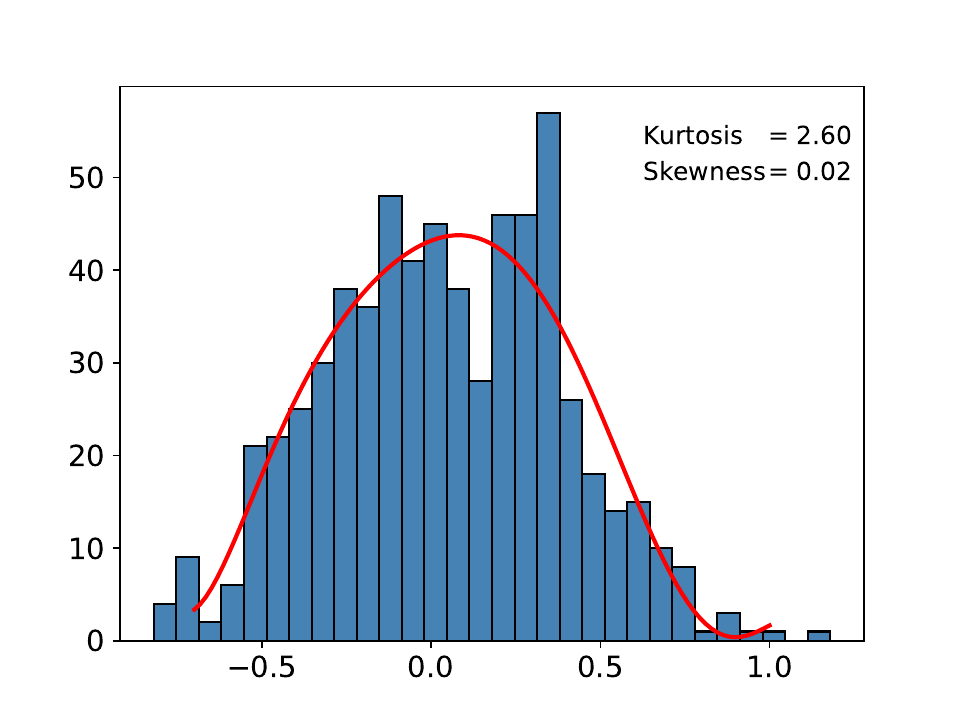} 
}
\subfigure[Kurtosis+Skewness Loss.] { \label{fig:our}
\includegraphics[width=0.492\columnwidth]{ 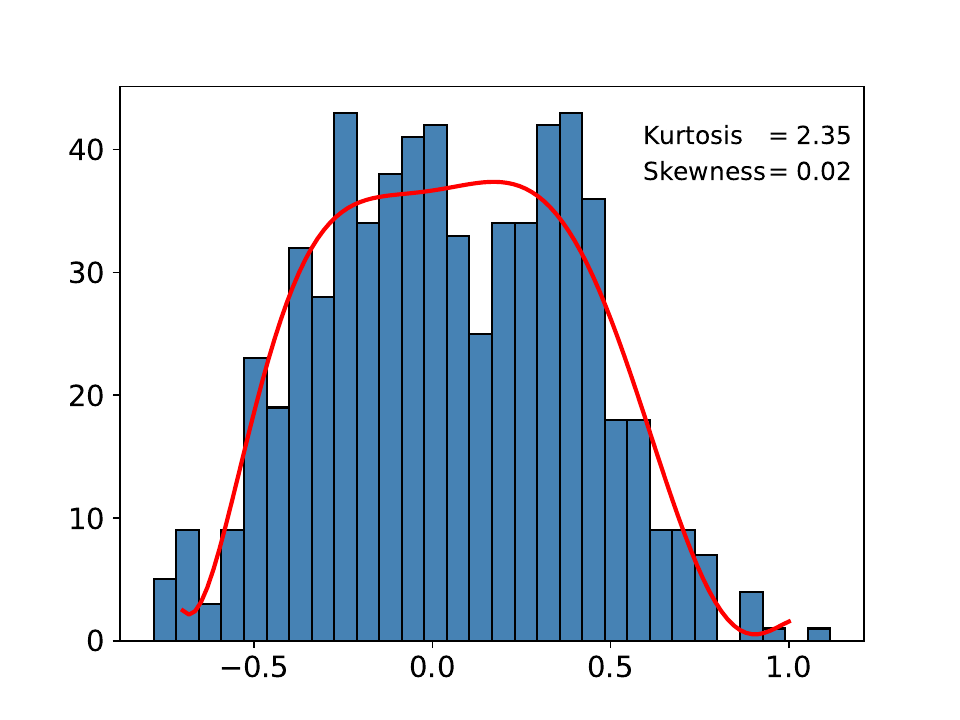} 
}
\caption{Ablation analysis of weights distribution from 21-th layer on elastic quantized ResNet20 with Kurtosis and Skewness regularization. The blue column represents the histogram distribution, and the red solid line represents the 7th order fitting curve of the data.}
\label{fig:wdr}
\end{figure*}

\paragraph{Effectiveness of Weight Distribution Regularization.}
To make the weight distribution of neural networks more suitable for elastic quantization, we introduce weight distribution regularization. 
Figure~\ref{fig:baseline} illustrates the weight distribution of the 21st layer of ResNet20 on the CIFAR10 dataset. The figure reveals that certain layers in ResNet architecture exhibit skewed and sharp distribution characteristics, as evidenced by the kurtosis value of 3.37 and the skewness value of 0.64.
The impact of such distribution phenomena on fixed-bit-width quantization is relatively insignificant. However, for elastic quantization with high robustness demands, such phenomena can significantly affect the overall performance, particularly for low bit widths.
Figure~\ref{fig:Kurtosis} and Figure~\ref{fig:Skewness} depict the effects of applying kurtosis and skewness regularization to the weights, respectively. Notably, Figure~\ref{fig:our} shows that simultaneously applying kurtosis and skewness regularization can lead to a distribution effect that is closer to uniform distribution, effectively eliminating data skewness and sharpness simultaneously.
Moreover, as presented in Table~\ref{tab:wdr}, incorporating kurtosis and skewness regularization can boost accuracy by nearly 1\% for the 2-bit scenario, while the average accuracy for 2, 4, and 8 bits can improve by 0.5\%.

\begin{table}[htbp]
  \centering
  \caption{Ablation study for weight distribution regularization.}
    \begin{tabular}{ccc}
    \hline
    ResNet20 & 2-bit & Avg 2-4-8-bit  \bigstrut\\
    \hline
    Baseline & 86.4\% & 90.3\% \bigstrut[t]\\
    + Kurtosis Loss & 87.3\% & 90.5\% \\
    + Skewness Loss & 86.9\% & 90.4\% \\
    Kurtosis+Skewness Loss & \textbf{87.3\%} & \textbf{90.7\%} \bigstrut[b]\\
    \hline
    \end{tabular}%
  \label{tab:wdr}%
\end{table}%

\begin{figure}
\centering
\includegraphics[width=1.0\columnwidth]{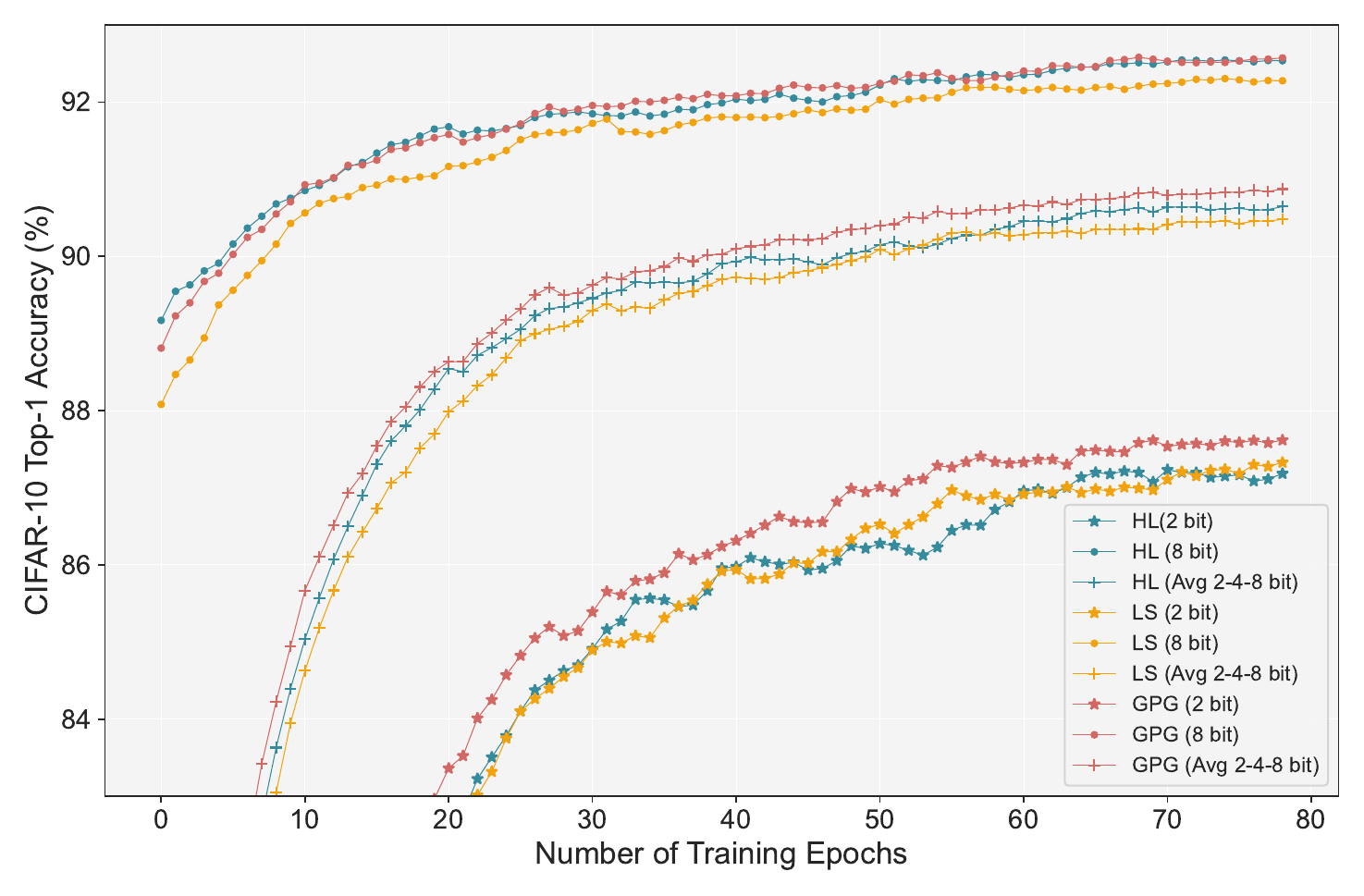}
\caption{Top-1 accuracy of ResNet20 on CIFAR-10 for different benchmarks (including 2bit, 8-bit, and 2-4-8 bit average accuracy). HL and LS denote hard label and label smoothing, respectively.} \label{fig:GPG}%
\end{figure}

\paragraph{Effectiveness of Group Progressive Guidance.}
In the training procedure of elastic quantization supernet, we adopt the training strategy of GPG proposed in Section~\ref{training strategy}. 
This strategy utilizes soft labels from the high bit-width subnet to progressively guide the low bit-width subnet, creating more coherence between the output of the high and low bit-width networks. As a result, the performance of the low bit-width subnet is substantially improved. 
The Convergence curve graph of ResNet20 trained using three different methods (hard label, label smoothing~\cite{LabelSmooth}, and our GPG method) on CIFAR-10 are presented in Figure~\ref{fig:GPG}. 
It can be observed that our proposed strategy consistently outperforms the other methods at 2 bit-width during training. Additionally, the performance for 2 bit-width is similar when using the label smoothing and hard label methods.
Furthermore, to demonstrate the training efficiency of the whole quantization supernet, we use the average precision of 2-4-8 bit-widths, and the average precision of our method is always the best.
When the bit-width is set to 8, although our GPG method is initially inferior to the hard label method during the first few epochs, our method steadily improves and is able to catch up with the hard label method, which demonstrates that our method can improve the accuracy of the low bit-width subnet without sacrificing the high bit-width performance.
 
\begin{table}[htbp]
  \centering
  \caption{Ablation study for learned vs. heuristic (min, mean, max) per-tensor quantization.}
    \begin{tabular}{cccc}
    \hline
    Per-channel & 2-bit & 4-bit & 8-bit \bigstrut[t]\\
    Baseline & 88.3\% & 91.9\% & 92.5\% \bigstrut[b]\\
    \hline
    Per-tensor & 2-bit & 4-bit & 8-bit \bigstrut[t]\\
    min   & 49.3\% & 72.7\% & 75.7\% \\
    mean  & 86.6\% & 91.8\% & 92.1\% \\
    max   & 87.0\% & 91.8\% & 92.2\% \\
    learnable & \textbf{87.2\%} & \textbf{92.4\%} & \textbf{92.5\%} \bigstrut[b]\\
    \hline
    \end{tabular}%
  \label{tab:pertensor}%
\end{table}%

\begin{figure*}[htbp]
\subfigure[ResNet18.] { \label{fig:corr-resnet18}
\includegraphics[width=0.65\columnwidth]{ 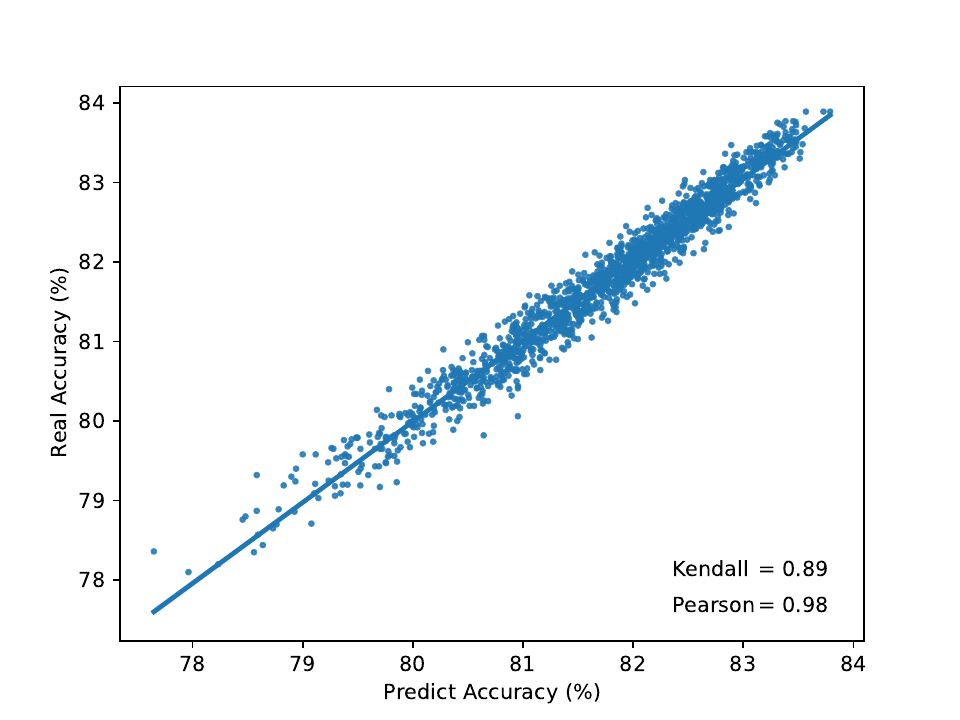}
}
\subfigure[MobileNetV2.] { \label{fig:corr-mobilenet_v2}
\includegraphics[width=0.67\columnwidth]{ 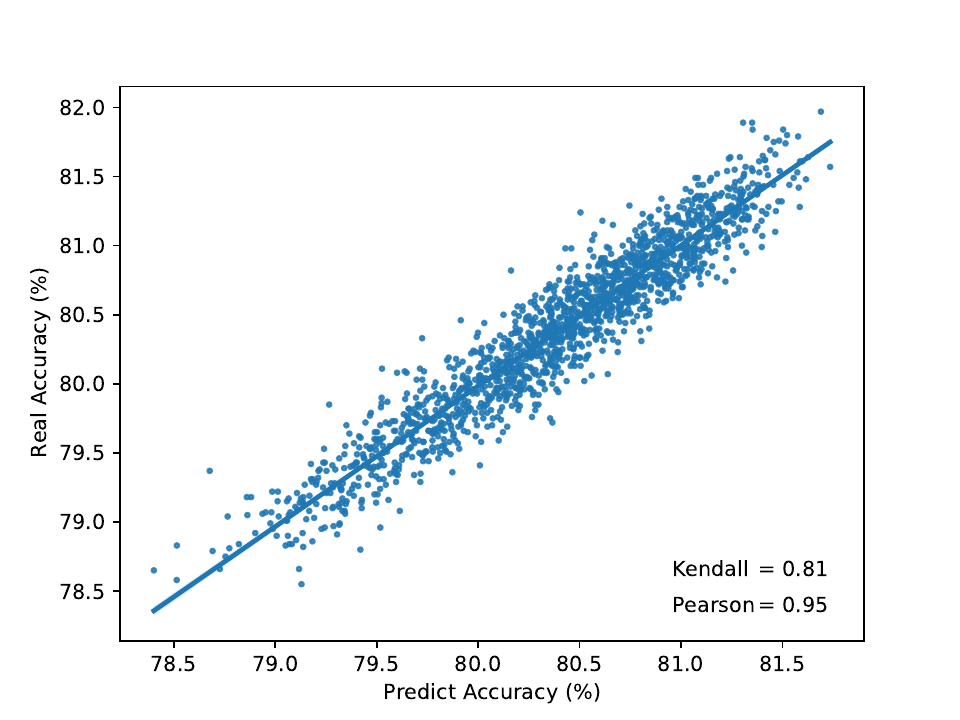}
}
\subfigure[EfficientNetB0.] { \label{fig:corr-efficientnet_b0}
\includegraphics[width=0.65\columnwidth]{ 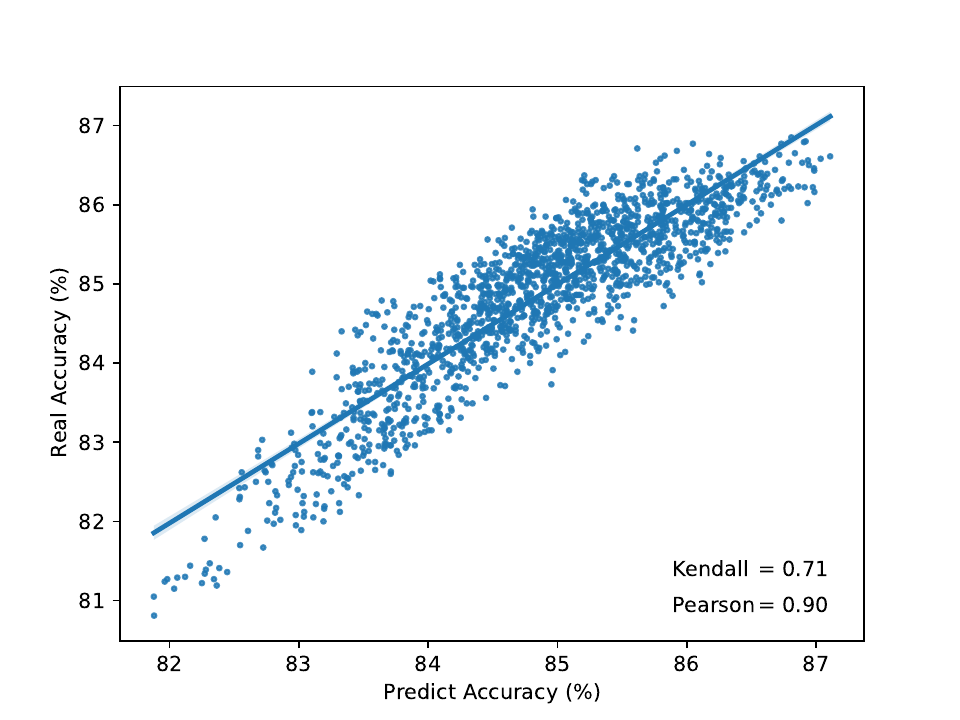}
}
\caption{Ablation analysis of CQAP Rank correlation between actual accuracy and predicted accuracy on split validation set of ImageNet.}
\label{fig:corr}
\end{figure*}

\paragraph{Learned vs. Heuristic Per-Tensor Quantization.}
Our proposed EQ-Net offers both per-channel and per-tensor quantization options. Per-channel quantization utilizes different step sizes for each convolution kernel, while per-tensor involves sharing a single step size across a layer of the network. Hence, exploring the efficacy of utilizing independent learnable parameters or heuristics on a per-channel basis for per-tensor quantization warrants investigation.
As shown in Table~\ref{tab:pertensor}, we compare the learnable method with three heuristic methods. The results demonstrate that the learnable method outperforms all three heuristics. Specifically, the learnable step size exhibits 0.2\%/0.6\%/0.3\% boosts over the best-performing heuristic method max at bit-widths of 2/4/8.
Among the three heuristics, the max achieves the highest accuracy, followed by the mean, which is only 0.4\%/0.1\% lower than the max at 2/8 bit-width, respectively. The worst performing method is min, which is approximately 20\% lower than the other two heuristics at any bit-width, this outcome is due to the narrow quantization value range that results from using the smallest step size, causing large quantization error.
Therefore, in our EQ-Net, we use independent learnable step sizes for per-tensor quantization.

\paragraph{Rank Preservation Analysis of Accuracy Predictor.}
As illustrated in Figure~\ref{fig:framework}, the mixed precision search can be conducted after the completion of quantization supernet training. 
During the search phase, we employ the CQAP, as proposed in Section~\ref{search}, as a proxy model for measuring accuracy.
Since CQAP is used to evaluate the performance of each mixed-precision model, it is imperative to guarantee a rank correlation between predictors and actual performance. We sampled 10k images from the training set of the ImageNet dataset and used the accuracy of this subset to measure the performance of the candidate subnet. In Figure~\ref{fig:corr}, we illustrate the rank correlation coefficients for three different supernets. It is evident that the Pearson coefficient is consistently above 0.90, and the Kendall coefficient is above 0.80 except for EfficientNetB0. It is demonstrated that there is a strong correlation between the predicted accuracy of our CQAP and the actual performance of the candidate subnet. The Kendall coefficient and Pearson coefficient for EfficientNetB0 are 0.71 and 0.90, respectively. These values are comparatively lower than those obtained for the other two networks under consideration. The reason for this slightly inferior performance can be attributed to the significant precision difference observed between symmetric and asymmetric quantization when applied to EfficientNetB0.

\section{Conclusion} \label{Conclusion}
In this paper, we have proposed Elastic Quantization Neural Networks (EQ-Net) that achieve hardware-friendly and efficient training through a one-shot weight-sharing quantization supernet. By training the supernet on designed elastic quantization space, EQ-Net can support subnets with both uniform and mixed-precision quantization without retraining. We propose two training schemes with Weight Distribution Regularization (WDR) and Group Progressive Guidance (GPG) techniques to optimize EQ-Net. We demonstrate that EQ-Net can achieve near-static quantization accuracy performance in an elastic quantization space.

\section*{Acknowledgments}
This work was supported in part by the National Natural Science Foundation of China (No. 62206003, No. 62276001, No. 62136008, No. U20A20306, No. U21A20512) and in part by the Excellent Youth Foundation of Anhui Provincial Colleges (No. 2022AH030013).

{\small
\bibliographystyle{ieee_fullname}
\bibliography{main}
}

\end{document}